\documentclass[sigconf]{acmart}

\usepackage{color, colortbl}
\definecolor{Gray}{gray}{0.9}
\usepackage{relsize}

\usepackage{multirow}

\newcommand{\RomanNumeralCaps}[1]
    {\MakeUppercase{\romannumeral #1}}
\usepackage{xcolor,colortbl}

\AtBeginDocument{%
  }

\copyrightyear{2023} 
\acmYear{2023} 
\setcopyright{acmlicensed}\acmConference[UrbanAI '23]{1st ACM SIGSPATIAL International Workshop on Advances in Urban-AI}{November 13, 2023}{Hamburg, Germany}
\acmBooktitle{1st ACM SIGSPATIAL International Workshop on Advances in Urban-AI (UrbanAI '23), November 13, 2023, Hamburg, Germany}
\acmPrice{15.00}
\acmDOI{10.1145/3615900.3628769}
\acmISBN{979-8-4007-0362-1/23/11}

\begin{document}

\title{CrashFormer: A Multimodal Architecture to Predict the Risk of Crash}


\author{Amin Karimi Monsefi}
\orcid{0000-0002-6101-2828}
\affiliation{%
  \institution{Ohio State University}
  \city{Columbus}
  \state{Ohio}
  \country{USA}
}
\email{karimimonsefi.1@osu.edu}

\author{Pouya Shiri}
\orcid{0000-0002-8037-9481}
\affiliation{%
  \institution{Circle Cardiovascular Imaging}
  \city{Calgary}
  \state{Alberta}
  \country{Canada}}
\email{pouya.shiri@circlecvi.com}

\author{Ahmad Mohammadshirazi}
\affiliation{%
  \institution{Ohio State University}
  \city{Columbus}
  \state{Ohio}
  \country{USA}
}
\email{mohammadshirazi.2@osu.edu}

\author{Nastaran Karimi Monsefi}
\affiliation{%
  \institution{Hamedan University of Technology}
  \state{Hamedan}
  \country{Iran}
}
\email{k.nastaran1998@gmail.com}

\author{Ron	Davies}
\affiliation{%
  \institution{Ohio State University}
  \city{Columbus}
  \state{Ohio}
  \country{USA}
}
\email{davies.404@osu.edu}

\author{Sobhan Moosavi}
\affiliation{%
  \institution{Zoox Inc.}
  \streetaddress{185 Berry Street}
  \city{San Francisco}
  \state{California}
\country{USA}
}
\email{smoosavi@zoox.com}

\author{Rajiv Ramnath}
\affiliation{%
  \institution{Ohio State University}
  \city{Columbus}
  \state{Ohio}
  \country{USA}
}
\email{ramnath.6@osu.edu}

\renewcommand{\shortauthors}{Monsefi et al.}

\begin{abstract}
Reducing traffic accidents is a crucial global public safety concern. Accident prediction is key to improving traffic safety, enabling proactive measures to be taken before a crash occurs, and informing safety policies, regulations, and targeted interventions. Despite numerous studies on accident prediction over the past decades, many have limitations in terms of generalizability, reproducibility, or feasibility for practical use due to input data or problem formulation. 
To address existing shortcomings, we propose \textbf{ \textit{CrashFormer}}, a multi-modal architecture that utilizes comprehensive (but relatively easy to obtain) inputs such as the history of accidents, weather information, map images, and demographic information. The model predicts the future risk of accidents on a reasonably acceptable cadence (i.e., every six hours) for a geographical location of $5.161$ square kilometers. CrashFormer is composed of five components: a sequential encoder to utilize historical accidents and weather data, an image encoder to use map imagery data, a raw data encoder to utilize demographic information, a feature fusion module for aggregating the encoded features, and a classifier that accepts the aggregated data and makes predictions accordingly. Results from extensive real-world experiments in 10 major US cities show that CrashFormer outperforms state-of-the-art sequential and non-sequential models by $1.8\%$ in F1-score on average when using ``sparse'' input data.
\end{abstract}

\ccsdesc[500]{Computing methodologies~Supervised learning by classification}
\ccsdesc[300]{Applied computing~Transportation}


\keywords{Accident Prediction, Long Sequence Time-Series Forecasting, Transformer, Multimodal Architecture}


\maketitle

\section{Introduction}
\label{sec:intro}



Traffic safety is a primary global public safety concern, with an estimated $1.35$ million in fatalities and $20-50$ million in injuries yearly due to road traffic crashes. These accidents result in loss of life and damage and have significant economic costs. Predictive models can be essential in improving traffic safety by identifying high-risk areas, informing safety policies and regulations, identifying patterns and trends in accident data, and developing targeted interventions to address specific factors contributing to accidents. \cite{world2015global}.

Given the importance of the problem, substantial research has been conducted in the field of accident risk prediction over the past few decades.

Previous studies have used various data sources to investigate and predict city-wide accident risk, including crash data, GPS data, road network attributes, land use features, and weather and population data. Examples of such studies include STCL-Net by Bao et al. \cite{bao2019spatiotemporal}, a crash risk prediction network by Pei Li et al. \cite{li2020real}, unsupervised clustering and XGBoost by Shi et al. \cite{Shi2019}, Deep Accident Prediction (DAP) by Moosavi et al. \cite{Moosavi2019}, RiskOracle framework by Zhou et al. \cite{Zhou2020}, a city-wide traffic accident risk forecasting model by Wang et al. \cite{wang2018cnn}, telematics data-based accident risk prediction by Hu et al. \cite{hu2019advancing}, and XSTNN, a Spatio-temporal Deep Learning solution by Medrano et al. \cite{de2021new}.

The existing studies have limitations such as being only applicable in specific locations, lacking generalizability (given their input data) \cite{lin2015novel, shangguan2021integrated}, insufficient accuracy in smaller areas \cite{yuan2018hetero, yuan2017predicting, de2021new}, dependence on a range of data attributes that may not be available in all regions \cite{najjar2017combining, Ren2018, yuan2018hetero, de2021new}, unsuitability for real-time applications \cite{Moosavi2019, de2021new, viswanath2021road, shi2021predicting, najjar2017combining}, or using over-simplified methods \cite{ihueze2018road, ren2018deep, lin2021intelligent}. These factors make them less practical for real-world application.

Incorporating additional information such as accident history, weather information, map images, and demographic data, is essential to a more precise prediction the risk of accidents. A more comprehensive view provides a better understanding of the various factors contributing to accidents \cite{eisenberg2004mixed, Moosavi2019}. Accident history offers insight into past incidents and helps identify high-risk areas \cite{jaroszweski2014influence, FARAHANI2023, tamerius2016precipitation}. Weather information, such as precipitation and temperature, can indicate how the weather conditions affect the risk of an accident \cite{Moosavi2019}. Map images offer a detailed view of road characteristics, and infrastructure \cite{monsefi2022will, najjar2017combining}. Demographic information such as population density, income levels, and age distribution help predict accident risk by identifying areas with high traffic congestion and the most at-risk groups \cite{jimenez2022identification}. Combining these data sources is crucial for developing effective safety measures, targeted actions, and policies to reduce accident risk \cite{mathew2022exploring, liu2022study, badrestani2019real, lebedev2022ndhmuw}.

To mitigate the existing shortcomings in the literature and utilize various valuable data sources, in this paper, we introduce \textbf{ \textit{CrashFormer}}, a multi-modal architecture for predicting the risk of accidents for each area. CrashFormer seeks to predict the risk of accidents in different regions by using a multi-source dataset, including the history of accidents, weather information, map images, and demographic information of each area. CrashFormer consists of five components: $(1)$ a sequential encoder, which takes time-series-based historical accident and weather data for a region, and its latent outputs contain encoded history data; $(2)$ an image encoder, which utilizes map image data for each area to encoder characteristics of roads for a feature vector; $(3)$ a raw feature encoder that uses demographic information of a region; $(4)$ a feature fusion module to aggregate all the encoded latent together and $(5)$  a classifier component that predicts the risk of accident for each location.

Our proposed approach centers on forecasting accident probabilities by utilizing historical accident and weather data presented in a sequential format. Additionally, we incorporate a comprehensive demographic dataset containing 150 attributes. This dataset was gathered through web scraping from an online portal (please refer to Section~\ref{Demographics} for detailed information about the database). Furthermore, we make use of map images corresponding to the locations of accident events. To encode data from these diverse sources, we employ two advanced models. Firstly, we employ the $FEDFormer$, a transformer-based model renowned for capturing long-range relationships within sequences, to encode historical data \cite{zhou2022fedformer}. Secondly, we utilize the Vision Attention Network ($VAN$) model to capture latent features within map images. The $VAN$ model facilitates self-adaptive and long-range correlations, allowing for effective analysis of visual information when predicting accident probabilities \cite{guo2022visual}.

In our experiments, we employ CrashFormer to predict accident risk (the probability of an accident) within a hexagonal area in the next 6 hours. We evaluate our proposal in 10 major cities in the United States, and the results show that our model outperforms the baselines. CrashFormer generally surpasses the next best network by $1.8\%$ improvement in the F1-score when predicting the risk of accidents. Additionally, we observe that the proposed model shows superior outcomes even when the training data is only available for certain parts of a city (referred to as \textit{spatial sparsity}). This implies that the model can generalize well to new areas without much effort to tweak model parameters.

\section{Related Work}
\label{sec:related}

Numerous previous studies have utilized historical crash data to predict the probability of traffic accidents. However, our research highlights that accident records from the past are not the sole significant factors; meteorological data and demographic information also play vital roles in anticipating accident risks. In this section, we present a brief overview of significant research efforts in this domain.

Bao et al. conducted a comprehensive investigation using a spatiotemporal convolutional LSTM network (STCL-Net) to analyze the risk of accidents across a city. They employed a multi-source dataset that included crash data, taxi GPS data, road network attributes, land use features, and weather and population data specifically for Manhattan in New York City \cite{bao2019spatiotemporal}. The study revealed that the predictive accuracy diminishes as the grid resolution increases. However, it is essential to note that since this study only uses data limited to a specific city, it may need further investigation to assess whether it is generalizable to locations with different accident patterns.


Shi et al. introduced a feature extraction and selection framework for accurate prediction of driver risk levels \cite{Shi2019}. By extracting approximately 1300 features related to driving behaviors, the authors employed unsupervised clustering to estimate vehicle risk. To address the problem of class imbalance, under-sampling techniques were applied to the less risky data. Furthermore, critical features were identified through feature importance ranking using XGBoost. This study demonstrated the effectiveness of this approach in accurately predicting vehicle risk levels.

Ren et al. collected significant traffic accident data and investigated the spatiotemporal correlation within traffic-related data \cite{Ren2018}. Their analysis revealed the non-uniform distribution of accident risks in time and space, exhibiting strong periodical temporal patterns and spatial correlations. The authors developed a Deep Neural Network (DNN) solution referred to as the Traffic Accident Risk Prediction Method (TARPML) to predict citywide crash risk. The reliance of this method on limited traffic accident data and its inability to predict risk at a fine-grained road level, are the shortcomings of this research.

Moosavi et al. proposed the Deep Accident Prediction (DAP) framework, which leverages multi-source data, including point-of-interest, weather data, and historical traffic events \cite{Moosavi2019}. The DAP framework incorporates recurrent, fully connected, and embedding components, and a publicly available large-scale accident dataset was utilized to evaluate its performance. The study demonstrated the efficacy of integrating traffic information, time, and points of interest in enhancing real-time risk prediction. However, the study takes a massive area for prediction, which helps to improve accuracy, but it is not useful in the real world, and the real world needs to predict accurately for a small area.

Zhou et al. introduced the RiskOracle framework for more granular crash risk prediction, down to a minute level \cite{Zhou2020}. They developed the Differential Time-varying Graph Neural Network (DTGN), which captures the dynamic nature of road networks and immediate changes in traffic conditions. To address biased risk predictions, region selection schemes were employed. The effectiveness of the proposed solution was verified using two real-world datasets.

Our proposed solution, CrashFormer, is an advanced time-series model designed to forecast the probability of accident risk in specific areas at any given time. This model is particularly advantageous for real-time applications due to its focus on a smaller geographic scale, specifically $5.161$ square kilometers, surpassing previous studies in terms of granularity. Furthermore, our approach introduces a novel integration of real-time traffic events, weather events, map images, and demographic information, which remains unexplored in existing research. Another notable advantage of our methodology is utilizing readily available and easily accessible data from public sources, eliminating the need for extensive and complex data collection procedures in other studies.

\section{Dataset}
\label{sec:dataset}
One of the important paper's contributions is offering a robust multi-source dataset for accident risk prediction. We gather data from multiple sources, use feature engineering to include additional useful features, and pre-process the data to better conform with the problem of concern i.e. predicting accident risks. Finally, we join and fuse the data from the different sources to create the dataset. The input datasets we use include information about accident history, weather information and events, demographic information for the accident area, and map images.

\subsection{Accident History}
To leverage the accident history, we start with the dataset proposed by Moosavi et al. \cite{moosavi2019countrywide} and add features to it. It comprises comprehensive data on approximately $6.2$ million accident records collected from Bing and MapQuest between $2016$ and $2021$ in the United States \footnote{download dataset from here https://smoosavi.org/datasets/us\_accidents}. These records provide detailed information about various accident attributes, including time, location, severity, duration, length of traffic impact, and descriptive details. We enhance this dataset by adopting the hexagon-based city partitioning technique proposed by Monsefi et al. \cite{monsefi2022will}, employing the H3 library developed by Uber \footnote{see https://github.com/uber/h3}.

Recognizing the significance of traffic patterns influenced by temporal factors, as highlighted by Badrestani et al. \cite{badrestani2019real}, we incorporated date and time considerations into our analysis. We introduced additional time-related features to capture the influence of the hour of the day, day of the week, and day of the month on traffic accident patterns. These features were extracted from the accident timestamps and appended to the dataset. Additionally, we determined the holiday status of the accident occurrence date and introduced an "isHoliday" flag to account for its potential impact on accident patterns.

Moreover, we considered the proximity of points of interest (POIs) to the accident locations. For this purpose, we utilized a one-hot vector of length 13, representing different types of POIs such as crossings, stations, amenities, bumps, giveways, traffic signals, no-exit points, railways, roundabouts, stops, traffic calming measures, junctions, and turning loops. This information provides additional context regarding the surroundings of the accident sites and contributes to a more comprehensive analysis of accident risk factors.

\subsection{Weather events}
To consider the weather conditions for the time accidents took place, we leverage a dataset collected by Moosavi et al.\cite{moosavi2019short}, containing fine-grained weather events recorded from 2016 to 2021, containing about $7.5$ million records. This dataset includes information about the type of weather event in addition to its severity in a specific weather station. The features are weather type (rain, snow, fog, etc.), time, duration, severity, and location.

\subsection{Demographics}
\label{Demographics}
Demographics are the socioeconomic statistics describing information about the population and its characteristics. Demographic analysis studies a population based on gender, age, and race. The data include income, education, marriage, death, birth, and employment. Socioeconomic information is a type of coarse-grained data that is included in this study. This data was gathered for about 45,000 US postal areas for different ZIP Codes. The collected demographic information includes four other groups of information for the residents of a specific postal area: demographics, housing, employment and income, and education. For data collection, we used data crawling strategies on an online portal \footnote{see https://www.unitedstateszipcodes.org/}. We collected 150 demographic features and created a public dataset \footnote{see the final dataset https://www.kaggle.com/datasets/aminkarimimonsefi/demographic-dataset}.

\subsection{Map Images}
\label{map_images}
Textural information about a specific area where accidents occur could help the model generate more accurate predictions. In this study, we have collected map images from OpenStreetMap\footnote{see https://tile.openstreetmap.org;} \cite{osm} using the coordinates of accident events. In other words, each accident is mapped to a unique area (hexagonal region). Afterward, the coordinates of the center of the areas are used to fetch a map image and assign it to the accident event. The location's zoning has been based on Uber's Hexagonal Hierarchical Spatial Indexing \footnote{see https://www.uber.com/blog/h3/} with a resolution of $R=7$, which corresponds to hexagonal areas with about $2604$ meters and $5161000$ square meters of edge length and area, respectively. Also, the zoom level of the OpenStreetMap is chosen to be $14$. We followed the method used by Monsefi et al. \cite{monsefi2022will} to collect the images.

\section{Research Question}
\label{sec:problem}

This section formally defines our research question by utilizing previously recorded traffic events, weather observations, and points of interest from \cite{Moosavi2019}. Additionally, we gather map images using the method outlined in \cite{monsefi2022will} and demographic information as outlined in Section~\ref{Demographics}. Given a set of accident events $\mathrm{C}$, we define a geographic region as follows.

\begin{definition}[Geographic Region]
    A geographical region $r$ is a \textit{hexagonal area}, as defined in Uber h3 library \citeN{sahr2003geodesic}. This library provides levels 0 to 15 with different areas \footnote{see \url{https://h3geo.org/docs/core-library/restable/}}. We have determined that a resolution level of $7$ is reasonable for our work. This resolution has an edge size of $2.604$ kilometers and an area of $5.161$ square kilometers, making it a useful moderate region size for various applications. Additionally, we gathered demographic information by using US zip codes. For each region $r$, we identified the closest zip code and assigned that information to the region.
\end{definition}

We use a database including high-resolution geographical images $\mathrm{M}$ represented as hexagonal tiles, as well as a dataset that contains the location of points of interest $\mathrm{P}$, such as stop signs, amenities, and traffic lights for a specific area. Using these datasets, we can now define our research question as follows. 

\vspace{3pt}
\noindent\textbf{Given:}
\begin{itemize}
    \item [--] A collection of geographical regions $\mathrm{R} = \{r_1, r_2, \dots, r_n\}$. 
    \item [--] A set of time intervals with a fixed duration $\mathrm{T} = \{t_1, t_2, \dots, t_m\}$, where $|t| = 6$ {\it hours}, for $t \in \mathrm{T}$.
    \item [--] A database of accident events $\mathrm{C}_r = \{c_1, c_2, \dots\}$ for $r \in \mathrm{R}$. 
    \item [--] A database of map image data $\mathrm{M}_r = \{m_1, m_2, \dots\}$ for $r \in \mathrm{R}$. 
    \item [--] A database of points-of-interest $\mathrm{P}_r = \{p_1, p_2, \dots\}$ for $r \in \mathrm{R}$. 
    \item [--] A single value $\mathrm{K}$ that shows length of sequenced data. 
\end{itemize}

\noindent\textbf{Create:}\vspace{-5pt}
\begin{itemize}
    \item [--] A representation $F_{rt}$ for region $r \in \mathrm{R}$ for a time interval $t \in \mathrm{T}$, using $P_r$, $C_r$ and $M_r$.
    \item [--] A binary label $L_{rt}$ for $F_{rt}$, where a value of $1$ indicates that the area has a high level of accident risk (if at least one accident was reported in that area) during $t$ in the region $r$; and $0$ otherwise. 
\end{itemize}

\noindent\textbf{Find:}\vspace{-5pt}
\begin{itemize}
    \item [--] A model $\mathcal{M}$ for predicting $L_{rt}$ using $\langle F_{rt_{i-K}}, F_{rt_{i-K-1}}, \dots, F_{rt_{i-1}} \rangle$. In other words, predict the label of the current time interval based on the observations from the $K$ recent intervals.
\end{itemize}

\noindent\textbf{Objective:}\vspace{-5pt}
\begin{itemize}
    \item [--] Minimize the prediction error.  
\end{itemize}


\section{Methodology}
\label{sec:method}

In this section, we describe a model called CrashFormer, which is used to predict the level of accident risk. The model takes in three types of input: a feature vector representation of accident events, map image data, and demographic data.

\begin{figure*}
    \centering  
    \includegraphics[width=\linewidth]
    {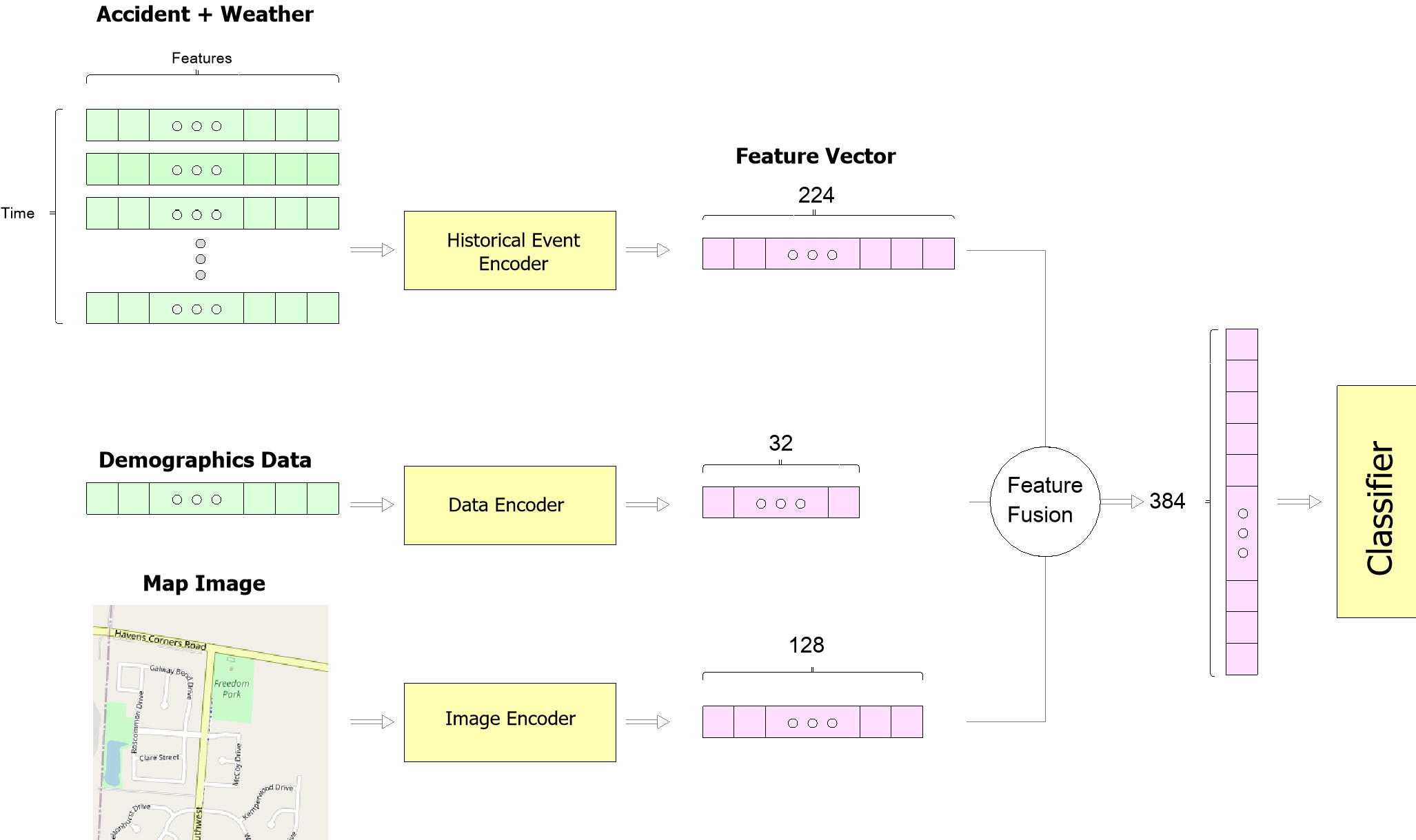}
    \caption{The architecture of CrashFormer. The sequential data based on accident and weather information, along with map images and demographic data, are each fed to a separate component, and the output feature vectors are concatenated. }
    \label{fig:crashformer}
\end{figure*}

\subsection{Sequential Feature Vector} 
\label{sec:feature-vector-representation}

We aggregate all the accident events in a geographical region and average them over a $t = 6\ \textit{hours}$ time interval and then create a feature vector representation. We consider the following features for the accident events:

\begin{itemize}
    \item \textbf{Weather}: We use a 8-dimensional vector to represent \textit{Weather Severity}, \textit{Precipitation}; and $6$ additional indicators representing special weather events \textit{Rain}, \textit{Fog}, \textit{Cold}, \textit{Snow}, \textit{Storm}, and \textit{Hail}. We use the weather data previously obtained by Moosavi et al. \citeN{moosavi2019short}.  
    
    \item \textbf{POI}: We create a vector of size $13$ for representing the existence of the point-of-interests (POIs, or map annotations) within a radius of $r$. The vector contains \textit{Amenity}, \textit{Traffic Calming}, \textit{Crossing},  \textit{Bump}, \textit{Junction}, \textit{Give Way}, \textit{No Exit}, \textit{Railway},  \textit{Station}, \textit{Stop},  \textit{Traffic Signal}, \textit{Roundabout} and \textit{Turning Loop}. The value for each item is $1$ if any of the accidents at time $t$ in area $r$ has a nearby point-of-interest; otherwise, it is $0$ \footnote{For more information about point-of-interests and their importance, see \cite{monsefi2022will}}.
    
    \item \textbf{Time-Related Contextual Information}: We construct a 4-dimensional vector for time information by aggregating all available data within a 6-hour window. From the accident timestamp, we extract the following information: \textit{Day of Week}, \textit{Month}, \textit{Day of Month}, and \textit{Is Holiday}. The use of this contextual time-related information should be useful, as it is believed that traffic patterns exhibit a cyclic variation based on hour, week, and month. To factor in the impact of work hours on traffic patterns, the hour of the day is computed from the accident timestamp, and the day of the week and month of the year are added as extra features.

    \item \textbf{Accident Information}: We create a 2-dimensional vector to represent the information related to the accident. We take the average of \textit{Accident Severity}, collected from \cite{Moosavi2019}. We also create a \textit{label} indicating whether an accident occurred in the region $r$ at time $t$.
\end{itemize}


For aggregating all the accidents that occurred during $t$ in a region $r$, we take the value for $POI$, weather type, and accident label data. For instance, if we have several accident events during the time window $t$ in area $r$, and if there is a specific POI near the location of one of the accidents, we set the corresponding $POI$ to $1$. 

\subsection{Map Image Representation }
\label{sec:map-image-representation}
Map images are a valuable source of contextual information in understanding the causes of accidents. Accidents may be less frequent in areas with a sparse road network and more frequent in urban areas with a dense road network. To capture this information, each geographical area is associated with an image obtained from $OpenStreetMap$ (OSM)\footnote{see \url{https://wiki.openstreetmap.org/wiki/Zoom_levels}} with a \textit{zoom\_level} of $14$. The map images used in the model are square-shaped and have a resolution of $256\times256$ pixels. Each pixel represents an area of $9.555$ meters. This means that the side size of the image is approximately $2.446$ km, and its total area is about $5.983 \textit{km}^2$. The map images can encompass the entire geographical region $r$, with an edge length of $2.604 \textit{km}$ and an area of $5.161 \textit{km}^2$. 


\subsection{Demographic Representation} 
The dataset of demographic data includes around $150$ features. We removed the columns containing descriptions and took $144$ features \footnote{for more information, see https://www.kaggle.com/datasets/aminkarimimonsefi/demographic-dataset}. In this dataset, we have some information related to area $r$ like \textit{Population}, \textit{Housing Units}, \textit{Density}, \textit{Gender}, \textit{Race}, etc. 

\subsection{CrashFormer}
\label{sec:CrashFormer}

\begin{table*}[ht!]
    \centering
    \setlength\tabcolsep{7pt}
    \caption{Number of hexagonal areas in each city}\vspace{-5pt}
    \begin{tabular}{|c|c|c|c|c|c|c|c|c|c|}
    \hline
   \rowcolor{Gray}
   \textbf{Houston} & \textbf{Miami}&
   \textbf{Los Angeles} & \textbf{Charlotte} & \textbf{Dallas} & \textbf{Austin} & \textbf{Atlanta}&  \textbf{Phoenix} & \textbf{Seattle} & \textbf{San Diego} \\
    \hline
        $386$  & $250$ & $133$ & $255$ & $210$ & $254$  & $185$ & $217$ & $105$ & $149$\\
    \hline
    \end{tabular}
    \label{tab:location_split_city}
\end{table*}

We present a multi-modal architecture named \textit{CrashFormer}, which leverages transformer-based architecture to forecast accident risk. This model incorporates a wide range of factors, including weather conditions, historical traffic data, the population density of the area, and road characteristics inferred from map images and demographic information. Building upon the findings of Najjar et al. \cite{najjar2017combining}, who demonstrated the feasibility of using satellite images to capture road safety similarity, we utilize map images to identify locations with similar levels of road safety based on environmental attributes, color variations (e.g., gray versus green), and the presence of specific objects (e.g., intersections, road types, roundabouts, and vegetation). By extracting visual features from map imagery, we effectively capture the characteristics and safety aspects of roads. To integrate these diverse data sources, we aggregate all relevant information within a specific location and time window, which serves as the input to the CrashFormer model. The proposed model comprises five components, each consisting of a deep neural network model. Figure \ref{fig:crashformer} illustrates the architecture of CrashFormer, and the subsequent sections provide- a detailed description of its constituent components.

\subsubsection{Historical Event Encoder}

We employ Frequency Enhanced Decomposed Transformer (FEDFormer), a state-of-the-art transformer-based network introduced by Zhou et al. \cite{zhou2022fedformer}, to extract the feature vector from the sequential weather and accident information. This network combines seasonal decomposition with a transformer to capture both overall trends and detailed structures, making it a highly precise network for time-series forecasting. FEDFormer has a linear complexity to the length of the time-series sequence, making it a more efficient solution than other transformers. This network also utilizes Fourier and Wavelet blocks for capturing time-series structures by mapping data to the frequency domain. For its lower computational and memory cost and its proven accuracy on several benchmarks, we chose FEDFormer as our feature extractor for the sequential data.

 The sequential data, as described in Section \ref{sec:feature-vector-representation}, is fed as input to the FEDFormer network. The input data is structured as $(B, L_S, N_F)$, where $B$ represents the batch size, $L_S$ indicates the sequence length, and $N_F$ denotes the length of the feature vector representation, which in our case is $27$. The output of this component is an encoded feature vector that encompasses valuable information about the historical records of traffic events and weather conditions. The resulting shape of the output is $(B, 224)$, signifying that the encoding effectively summarizes the entire historical information into a vector of size $224$.

\subsubsection{Image Encoder}

We utilize VAN to extract useful road network information from map images. VAN is an attention-based model introduced by Guo et al. \cite{guo2022visual}. We used $VAN-B1$ for our feature encoding. A critical attribute of attention techniques is adaptively modifying the representation based on the input characteristic. This helps us encode the map images. This module takes images with a shape of $(B, C, H, W)$ as input, where $B$ is the batch size, $C$ is the number of image channels $(3)$, and $H$ and $W$ are the image height and width, respectively (see Section \ref{sec:map-image-representation} for details on images). The output of this component is a vector of size $(B, 128)$.

\subsubsection{Data Encoder}
This component is a simple multi-layer feed-forward sub-network to encode the zone-wide demographic information (explained in Section \ref{Demographics}). We selected and normalized 144 demographic features and used them as input to this sub-network, which encodes the information into a small vector. The output vector size is $(B, 28)$ ($B$ is data batch size).

\subsubsection{Feature Fusion} The process of consolidating latent features extracted from various sources involves amalgamating these features and generating a novel latent feature representation. This newly formed latent feature is subsequently input into the classifier layer for further analysis and classification. This approach facilitates the integration of diverse information streams, enhancing the model's ability to make informed decisions.

\subsubsection{Classifier}
The final component of our model is a simple multi-layer feed-forward network. We combined the outputs of previous components into a single vector of size $(B, 380)$ with a Feature fusion layer and input it into this sub-network to determine the accident risk for area $r$ at time $t$. By using $softmax$ as the activation function in this component, we are able to make the risk prediction.

\section{Experiment and Results}
\label{sec:experiment}

\begin{table*}[]
    \centering
    \setlength\tabcolsep{7pt}
\caption{Evaluation of CrashFormer with varying length of sequences, $F1\_1$ and $F1\_0$ means $F1\_score$ for label one and zero }\vspace{-5pt}
\begin{tabular}{|c|cc|cc|cc|cc|}
\hline
\rowcolor{Gray}
Sequence & \multicolumn{2}{c|}{len=4} & \multicolumn{2}{c|}{len=8} & \multicolumn{2}{c|}{len=12} & \multicolumn{2}{c|}{len=16} \\ \hline
\rowcolor{Gray}
Metrics & \multicolumn{1}{c|}{F1\_1} & F1\_0 & \multicolumn{1}{c|}{F1\_1} & F1\_0 & \multicolumn{1}{c|}{F1\_1} & F1\_0 & \multicolumn{1}{c|}{F1\_1} & F1\_0 \\ \hline
Houston & \multicolumn{1}{c|}{\textbf{0.6539}} & \textbf{0.9808} & \multicolumn{1}{c|}{0.6391} & 0.9793 & \multicolumn{1}{c|}{0.6286} & 0.9787 & \multicolumn{1}{c|}{0.5961} & 0.9757 \\ \hline
Seattle & \multicolumn{1}{c|}{0.5058} & 0.9717 & \multicolumn{1}{c|}{\textbf{0.5093}} & \textbf{0.9721} & \multicolumn{1}{c|}{0.4996} & 0.9712 & \multicolumn{1}{c|}{0.4942} & 0.9706 \\ \hline
Miami & \multicolumn{1}{c|}{\textbf{0.5822}} & \textbf{0.9770} & \multicolumn{1}{c|}{0.5609} & 0.9750 & \multicolumn{1}{c|}{0.5377} & 0.9727 & \multicolumn{1}{c|}{0.5217} & 0.9711 \\ \hline
Los Angeles & \multicolumn{1}{c|}{\textbf{0.6698}} & \textbf{0.9515} & \multicolumn{1}{c|}{0.6462} & 0.9469 & \multicolumn{1}{c|}{0.6297} & 0.9435 & \multicolumn{1}{c|}{0.6136} & 0.9401 \\ \hline
Charlotte & \multicolumn{1}{c|}{\textbf{0.6167}} & \textbf{0.9798} & \multicolumn{1}{c|}{0.6015} & 0.978 & \multicolumn{1}{c|}{0.5935} & 0.9781 & \multicolumn{1}{c|}{0.5818} & 0.9772 \\ \hline
Dallas & \multicolumn{1}{c|}{\textbf{0.6156}} & \textbf{0.9792} & \multicolumn{1}{c|}{0.6034} & 0.9782 & \multicolumn{1}{c|}{0.5929} & 0.9773 & \multicolumn{1}{c|}{0.5845} & 0.9767 \\ \hline
Austin & \multicolumn{1}{c|}{\textbf{0.5547}} & \textbf{0.9829} & \multicolumn{1}{c|}{0.5444} & 0.9823 & \multicolumn{1}{c|}{0.5331} & 0.9816 & \multicolumn{1}{c|}{0.5213} & 0.9808 \\ \hline
Atlanta & \multicolumn{1}{c|}{0.5509} & 0.9820 & \multicolumn{1}{c|}{\textbf{0.5526}} & \textbf{0.9821} & \multicolumn{1}{c|}{0.5496} & 0.9819 & \multicolumn{1}{c|}{0.5427} & 0.9814 \\ \hline
Phoenix & \multicolumn{1}{c|}{0.4993} & 0.9814 & \multicolumn{1}{c|}{\textbf{0.5001}} & \textbf{0.9815} & \multicolumn{1}{c|}{0.4943} & 0.9811 & \multicolumn{1}{c|}{0.4857} & 0.9805 \\ \hline
San Diego & \multicolumn{1}{c|}{\textbf{0.5507}} & \textbf{0.9802} & \multicolumn{1}{c|}{0.5496} & 0.9801 & \multicolumn{1}{c|}{0.5485} & 0.9801 & \multicolumn{1}{c|}{0.5451} & 0.9798 \\ \hline  \hline 

Average & \multicolumn{1}{c|}{\textbf{0.57996}} & \textbf{0.97665} & \multicolumn{1}{c|}{0.57071} & 0.97555 & \multicolumn{1}{c|}{0.56075} &  0.97462 & \multicolumn{1}{c|}{0.54867} & 0.97339 \\ \hline
\end{tabular}

\label{tab:Scenario_1}
\end{table*}

In this section, we analyze the performance of $CrashFormer$ in comparison to other leading models and baselines. We conducted our experiments using data on accidents and weather from $June$ $2016$ to $December$ $2021$ for $10$ major cities in the United States under \textit{four} different scenarios (as outlined in subsequent sections). 

\subsection{Experiment Setup}
We implemented $CrashFormer$ in $Python$ and used $Pytorch$ \citeN{NEURIPS2019_9015} and $scikit-learn$ \citeN{JMLR:v12:pedregosa11a} packages for implementing our model and baseline models. We used Ohio Supercomputer Center \citeN{OhioSupercomputerCenter1987} machines for running our experiments.

We chose \textit{Adam} optimizer to train $CrashFormer$, and selected $200$ as the maximum number of epochs for training. To avoid overfitting, we applied an early stopping policy, where training will stop if the value of the loss on the validation set does not decrease after $10$ consecutive epochs. The initial learning rate was set to $10^{-3}$, and if no improvements could be observed for $5$ consecutive epochs, then the learning rate is reduced by a factor of $0.9$ (i.e., $\textit{LR\_{new}} = \textit{LR} \times 0.9$). This reduction could potentially continue until the learning rate reaches $10^{-6}$. 
We used \textit{Binary Cross Entropy} for the loss function because finding a level of accident risk for different areas can be modeled as a binary classification problem. This loss function has been proven to work effectively for this class of problems. 

\subsection{Data Description}
We trained and validated our $CrashFormer$ in ten cities (see table \ref{tab:location_split_city}). The choice of these cities was primarily to achieve diversity in traffic records (accident) and weather conditions, population, population density, and urban characteristics (road network, the prevalence of urban versus highway roads, etc.).

As explained in Section \ref{sec:CrashFormer}, we prepared and pre-processed data to use in the machine learning process. Each data record includes aggregated feature vector representation for a $6$ hours time interval for a specific area $r$, and we used demographic data as raw data for area $r$. Also, we used a map image for area $r$ to represent the corresponding geographical area. The goal is to predict the level of accident risk, that is, a binary classification problem. We split the cities into small areas of similar size using the $H3$ library. Table~\ref{tab:location_split_city} shows the number of regions in each city. Houston in Texas, with $386$ regions, is the largest city among the selected cities. In addition, splitting time into $t=6$ hours means that we have four samples per day. Therefore, we have $8080$ time windows for each $H3$ area over the entire dataset. As one could expect, the data suffers a significant class imbalance issue, where samples with label $1$ (i.e., high level of accident risk) are quite rare. For a total of $17,517,440$ samples, there are only $644,322$ records with a label $1$ (i.e., $3.67\%$ of the data). To mitigate this issue, we weighed the classes in the loss function and empirically found weights $15.327$ and $0.516$ for labels $1$ and $0$, respectively. 

\begin{table*}[]
    \centering
    \setlength\tabcolsep{7pt}
\caption{Effect of map images and demographic information on the risk of accident for that area. wo/Img means without map image. wo/Demog means without Demographic} \vspace{-5pt}
\begin{tabular}{|c|cc|cc|cc|cc|}
\hline
\rowcolor{Gray}
Model & \multicolumn{2}{c|}{CrashFormer}  & \multicolumn{2}{c|}{CrashFormer + wo/Img} & \multicolumn{2}{c|}{CrashFormer + wo/Demog} & \multicolumn{2}{c|}{CrashFormer wo/Img + wo/Demog} \\ 

\hline
\rowcolor{Gray}
Metrics & \multicolumn{1}{c|}{F1\_1} & F1\_0 & \multicolumn{1}{c|}{F1\_1} & F1\_0 & \multicolumn{1}{c|}{F1\_1} & F1\_0 & \multicolumn{1}{c|}{F1\_1} & F1\_0 \\ \hline
Houston & \multicolumn{1}{c|}{\textbf{0.6539}} & 0.9808 & \multicolumn{1}{c|}{0.6446} & 0.9812 & \multicolumn{1}{c|}{0.6498} & \textbf{0.9817} & \multicolumn{1}{c|}{0.6398} & 0.9813 \\ \hline
Seattle & \multicolumn{1}{c|}{\textbf{0.5058}} & \textbf{0.9717} & \multicolumn{1}{c|}{0.4886} & 0.9706 & \multicolumn{1}{c|}{0.4941} & 0.9706 & \multicolumn{1}{c|}{0.4831} & 0.9695 \\ \hline
Miami & \multicolumn{1}{c|}{\textbf{0.5822}} & 0.9770 & \multicolumn{1}{c|}{0.5627} & 0.9780 & \multicolumn{1}{c|}{0.5797} & \textbf{0.9789} & \multicolumn{1}{c|}{0.5604} & 0.9780 \\ \hline
Los Angeles & \multicolumn{1}{c|}{\textbf{0.6698}} & 0.9515 & \multicolumn{1}{c|}{0.6546} & 0.9646 & \multicolumn{1}{c|}{0.6614} & \textbf{0.9654} & \multicolumn{1}{c|}{0.6683} & 0.9653 \\ \hline
Charlotte & \multicolumn{1}{c|}{0.6167} & \textbf{0.9798} & \multicolumn{1}{c|}{0.6091} & 0.9746 & \multicolumn{1}{c|}{\textbf{0.6173}} & 0.9750 & \multicolumn{1}{c|}{0.6072} & 0.9742 \\ \hline
Dallas & \multicolumn{1}{c|}{\textbf{0.6156}} & \textbf{0.9792} & \multicolumn{1}{c|}{0.6141} & 0.9713 & \multicolumn{1}{c|}{0.5938} & 0.9689 & \multicolumn{1}{c|}{0.5912} & 0.9684 \\ \hline
Austin & \multicolumn{1}{c|}{\textbf{0.5547}} & \textbf{0.9829} & \multicolumn{1}{c|}{0.5370} & 0.9749 & \multicolumn{1}{c|}{0.5424} & 0.9756 & \multicolumn{1}{c|}{0.5523} & 0.9764 \\ \hline
Atlanta & \multicolumn{1}{c|}{0.5509} & 0.9820 & \multicolumn{1}{c|}{0.5550} & 0.9653 & \multicolumn{1}{c|}{\textbf{0.5572}} & 0.9814 & \multicolumn{1}{c|}{0.5594} & \textbf{0.9822} \\ \hline
Phoenix & \multicolumn{1}{c|}{0.4993} & \textbf{0.9814} & \multicolumn{1}{c|}{0.4988} & 0.9808 & \multicolumn{1}{c|}{\textbf{0.4998}} & 0.9808 & \multicolumn{1}{c|}{0.4938} & 0.9804 \\ \hline
San Diego & \multicolumn{1}{c|}{\textbf{0.5507}} & 0.9802 & \multicolumn{1}{c|}{0.5195} & 0.9804 & \multicolumn{1}{c|}{0.5370} & \textbf{0.9813} & \multicolumn{1}{c|}{0.5259} & 0.9806 \\ \hline  \hline

Average & \multicolumn{1}{c|}{\textbf{0.57996}} & \textbf{0.97665} & \multicolumn{1}{c|}{0.56858} & 0.97417 & \multicolumn{1}{c|}{0.57325} & 0.97596 & \multicolumn{1}{c|}{0.56819} & 0.97573 \\ \hline
\end{tabular}
\label{tab:Scenario_2}
\end{table*}

\subsection{Baseline Models}


To gauge the effectiveness of our proposed solution, we carried out a thorough evaluation encompassing both Transformer-based and non-Transformer-based models. These models represent the cutting edge in time series modeling. Our aim was to assess how our approach performs in comparison to existing methods.

\begin{itemize}

\item \textbf{Transformer-based models}
\begin{itemize}
    \item \textbf{Transformer}: Vaswani et al. introduced $Transformer$ \cite{vaswani2017attention}. Recently, Transformer-based solutions have seen an increase in long-term time series forecasting use. The transformer we used for evaluation has full attention on the encoder and decoder parts. Also, we have two layers at the encoder part and one at the decoder part.
    
    \item \textbf{Informer}: The Informer model is a deep learning model that captures long-range dependencies from long sequences of time-series data \cite{zhou2021informer}. We used this model with the exact settings used in \cite{zhou2021informer}.
\end{itemize}

\item \textbf{Non-Transformer-based models}

\begin{itemize}
   \item \textbf{DLinear}: Recently, Zeng et al. \cite{zeng2022transformers} introduced one-layer linear models referred to as $LTSF-Linear$ and compare their models to transformer-based solutions to forecast long-term time series data. We used their model with their recommended settings to assess its performance on our problem.

    
\end{itemize}

\end{itemize}

Since some baseline models are not designed to work with sequential data, the input data is vectorized before feeding these models. The vectorization process uses accident and weather events data and converts the time series data into a single vector. This is necessary as the baseline models cannot process sequential data in their original form and must be transformed into a feature vector for input.

\subsection{Results and Discussions}

In this section, we conduct four experiments to evaluate the performance and generalization abilities of our proposed model, CrashFormer. The first experiment focuses on investigating the impact of hyperparameter selection on the model's performance. We carefully tune the hyperparameters and analyze their effect on the prediction results. In the second experiment, we explore the enhancement in performance achieved by incorporating map images and demographic information as additional features. By integrating these data sources, we aim to capture more comprehensive contextual information that can improve the accuracy of accident risk predictions. The third and fourth experiments involve different data-splitting approaches to further assess CrashFormer's performance. In the third experiment, we employ a training set consisting of data up until December 30, 2020, while using the remaining data as the test set. This scenario assumes the complete availability of accident data for all areas and allows us to evaluate the model's performance under ideal conditions. In the fourth experiment, we simulate a real-world scenario of spatial sparsity by randomly selecting certain areas within a city for the training set while using the remaining data as the test set. This setup reflects the challenge of making accurate predictions for an entire city when only limited data from specific areas are available. By examining CrashFormer's performance in this context, we gain insights into its ability to generalize effectively and make reliable predictions with limited data.

\begin{figure*}
    \centering  \includegraphics[scale=0.5]{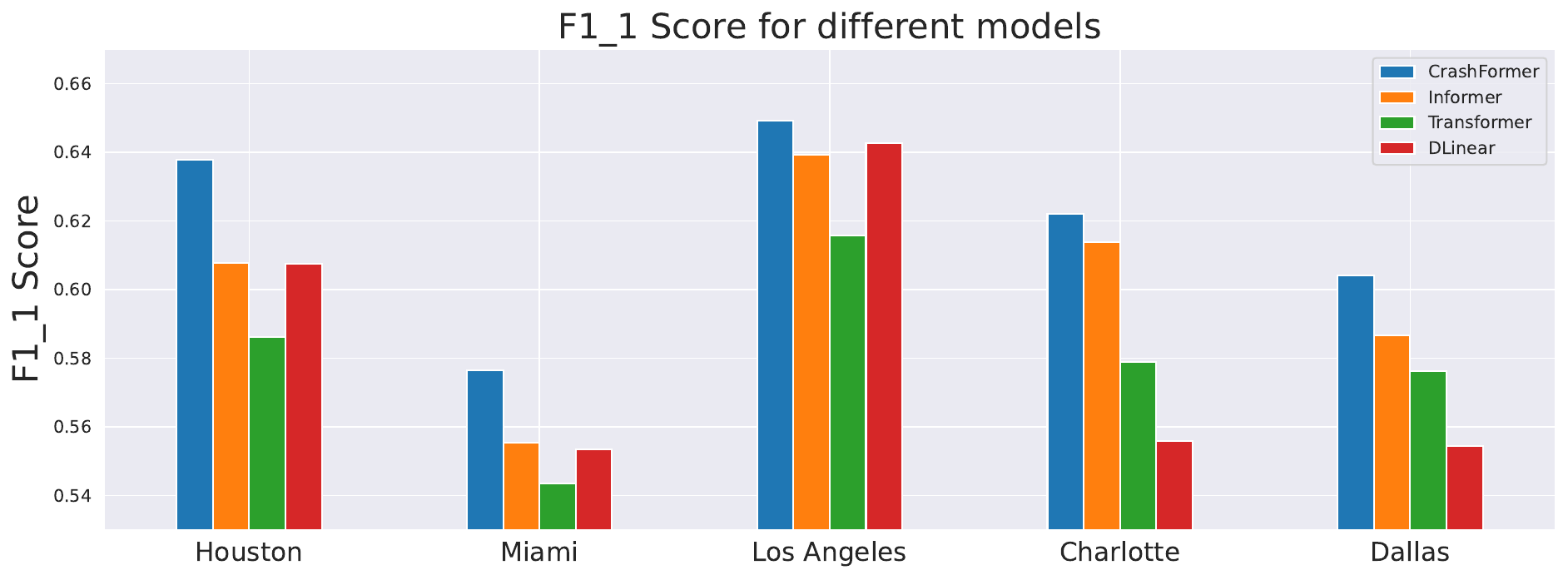}
    \caption{Comparing $CrashFormer$ to baselines. $F1\_1$ denotes $F1\_score$ for label one (high accident risk).}
    \label{fig:compare_baseline}
\end{figure*}

\subsubsection{Experiment \RomanNumeralCaps{1}}
\label{res:Scenario1}
In Experiment \RomanNumeralCaps{1}, the primary objective is to identify the optimal value for the hyperparameters of CrashFormer, with a particular focus on the length of sequential data, denoted as $L_S$. This experiment plays a crucial role in determining the configuration that yields the highest prediction performance.

Table \ref{tab:Scenario_1} presents the experimental results for various choices of $L_S$, specifically ranging from 4 to 16. With $L_S=4$, the data is divided into four consecutive time windows, each spanning a duration of 6 hours. Consequently, CrashFormer incorporates information from the past 24 hours to predict the risk of accidents for the subsequent 6-hour period. The evaluation metric employed in this experiment is the $F1\_Score$, which takes into account both label one (high accident risk) and label zero (low accident risk) to account for the class imbalance present in the dataset. The $F1\_1$ and $F1\_0$ scores represent the $F1\_Score$ calculated specifically for label one and label zero, respectively.

To conduct this experiment, the dataset is partitioned based on regions (or areas) and time intervals. The training set comprises $70\%$ of the data, while $10\%$ is allocated for validation, and the remaining $20\%$ is utilized as the test set.

Analyzing the results presented in Table \ref{tab:Scenario_1}, it is evident that in the majority of cases, utilizing data from four previous consecutive time windows ($L_S=4$) yields superior prediction performance compared to the other choices. Across all ten cities, averaging the results shows that selecting $L_S=4$ leads to a substantial improvement in the $F1\_1$ score. Specifically, compared to $L_S=8$, $L_S=12$, and $L_S=16$ configurations, adopting $L_S=4$ demonstrates enhancements of $1.62\%$,$ 3.42\%$, and $5.70\%$ in the $F1\_1$ score, respectively.

These findings provide strong evidence that CrashFormer achieves its best prediction performance when incorporating data from the past 24 hours (represented by $L_S=4$). By leveraging this time window, the model effectively captures the relevant temporal patterns and dependencies necessary for accurate accident risk prediction. As a result, for the subsequent experiments, the choice of $L_S=4$ is adopted as the standard setting.

\subsubsection{Experiment \RomanNumeralCaps{2}}
In Experiment \RomanNumeralCaps{2}, the aim is to assess the impact of incorporating map images and demographic data, in addition to sequential data, on the performance of CrashFormer. This experiment serves as an ablation study to understand the individual contributions and combined effect of these additional data sources.

To conduct the experiment, the model's input is modified by excluding map images, demographic data, or both. The predictive performance of CrashFormer is then evaluated under each configuration. Similar to Experiment \RomanNumeralCaps{1}, the data is split based on regions (or areas) and time intervals using the same strategy.

Table \ref{tab:Scenario_2} presents the results obtained from this experiment, showcasing the impact of each data source on CrashFormer's performance. Across most cities, the inclusion of both demographic and map images yields improvements in the model's prediction performance. Averaging the results obtained from all ten cities, it is observed that excluding map images leads to a reduction of $2.0\%$ in the $F1\_1$ score, excluding demographic data results in a $1.17\%$ reduction, and excluding both (utilizing only sequential data) leads to a $2.07\%$ reduction.

These findings highlight the importance of incorporating map images and demographic information in CrashFormer to enhance its predictive capabilities. The addition of map images, in particular, has a more pronounced impact on the model's predictions compared to demographic data. This suggests that spatial information captured through map imagery plays a significant role in identifying accident risk, enabling CrashFormer to better understand the geographical context and identify potential accident-prone areas.

The observed reductions in the $F1\_1$ score when excluding specific data sources underscore the competitive advantage of CrashFormer in leveraging a comprehensive set of inputs. By incorporating map images and demographic data alongside sequential information, CrashFormer demonstrates its ability to capture diverse factors that contribute to accident risk, resulting in improved prediction performance.

\subsubsection{Experiment \RomanNumeralCaps{3}}
In Experiment \RomanNumeralCaps{3}, the objective is to evaluate the performance of CrashFormer in predicting accident risk after a certain date, assuming complete availability of data for all areas. The dataset is split into training and test sets based on time, with the training set consisting of data from June 21, 2016, to Dec 30, 2020, and the test set containing the remaining data.

To assess the effectiveness of CrashFormer in comparison to state-of-the-art models for time series prediction, several baselines are utilized. These baselines include the Informer and DLinear. Notably, the input to all baselines in this experiment is limited to sequential data only, without the inclusion of map images or demographic information.

Figure \ref{fig:compare_baseline} presents the results of this experiment, specifically focusing on the $F1\_1$ scores of CrashFormer compared to the different baselines for the five selected cities: Houston, Miami, Los Angeles, Charlotte, and Dallas. The comparison demonstrates the superiority of CrashFormer over the top-performing baseline models in terms of predictive accuracy.

For Houston, CrashFormer outperforms the best baseline model by $4.73\%$ in terms of $F1\_1$ score. Similarly, for Miami, CrashFormer achieves a $3.64\%$ improvement over the top baseline. In the case of Los Angeles and Dallas, CrashFormer surpasses the best baseline by $1.00\%$ and $2.73\%$, respectively.

The exceptional performance of CrashFormer can be attributed to several factors. Firstly, the utilization of a more advanced model architecture, namely FEDFormer, contributes to its superior predictive capabilities. FEDFormer incorporates both sequential data and additional map images and demographic information, enabling the model to capture a broader range of features and contextual factors that influence accident risk.

The inclusion of demographic and map data in CrashFormer proves to be advantageous, as it provides valuable insights into the underlying factors influencing accident occurrence. By incorporating demographic information, CrashFormer can consider population density, age distribution, and other relevant demographic factors that may impact accident risk. The integration of map images allows the model to analyze the spatial characteristics of an area, identifying road network structures, intersections, and other geographical elements that contribute to accident proneness.

\subsubsection{Experiment \RomanNumeralCaps{4}}
In Experiment \RomanNumeralCaps{4}, the objective is to evaluate the impact of spatial sparsity on the prediction performance of CrashFormer. The experiment focuses on the city of Houston, which has the highest number of areas or hexagonal zones in the dataset. The dataset is divided into training, validation, and testing sets using a similar strategy as described in Section \ref{res:Scenario1}.

For this experiment, the performance of CrashFormer is compared to the baselines in terms of $F1\_1$ score. The baselines employ sequential data as input, while CrashFormer utilizes data from all three sources: sequential data, map images, and demographic data. The purpose is to assess how CrashFormer's comprehensive approach performs when faced with spatial sparsity.

\begin{figure}
    \centering \includegraphics[scale=0.5]{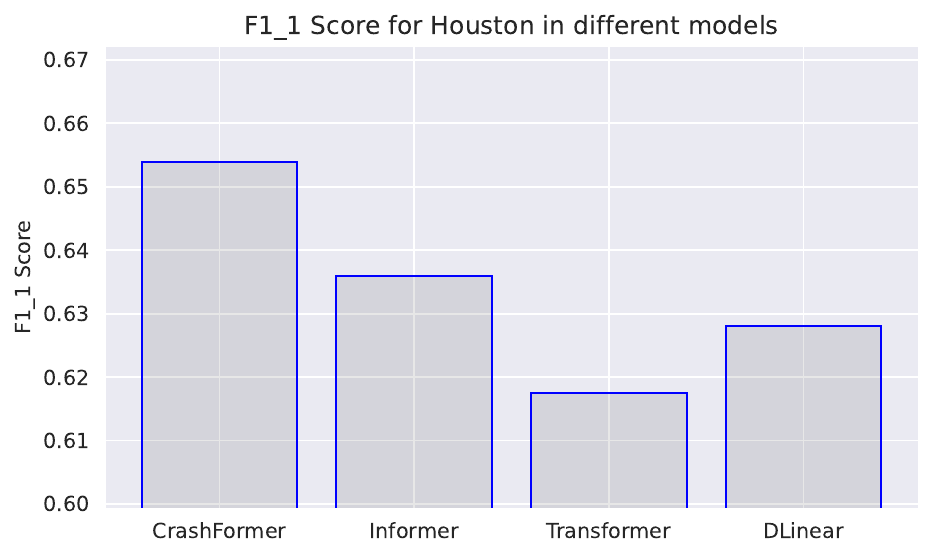}
    \caption{Comparing $CrashFormer$' with the baselines to test the impact of spatial sparsity on accident prediction in Houston (TX)}
    \label{fig:compare_baseline_scen4}
\end{figure}

Figure \ref{fig:compare_baseline_scen4} presents the results of this experiment, highlighting the superiority of CrashFormer over the baselines. With an $F1\_1$ score of $0.6539$, CrashFormer outperforms all other baselines for the city of Houston. Specifically, CrashFormer achieves a $2.737\%$ improvement in $F1\_1$ score compared to the best-performing baseline model, Informer.

The enhanced performance of CrashFormer can be attributed to multiple factors. Firstly, the inclusion of demographic data and map images provides valuable additional information for modeling the spatial characteristics of accident risk. The incorporation of demographic data allows CrashFormer to consider the demographic profile of different areas, such as population density, age distribution, and other socio-economic factors, which can influence accident occurrence.

Furthermore, the integration of map images enables CrashFormer to analyze the road network structures, intersections, and other geographical features that contribute to accident proneness. This spatial awareness allows CrashFormer to capture the spatial dependencies and patterns in accident data, resulting in improved prediction accuracy even in sparsely sampled areas.

Moreover, the use of the FEDFormer model architecture further enhances CrashFormer's performance. FEDFormer effectively integrates the sequential, map, and demographic data, enabling the model to capture complex temporal and spatial relationships for accurate accident risk prediction. The combination of these factors allows CrashFormer to generalize well to new areas and make better predictions for areas where the network has not been previously trained.

The results of Experiment \RomanNumeralCaps{4} demonstrate the superior generalization and predictive capabilities of CrashFormer when faced with spatially sparse data. By surpassing the performance of the baselines, CrashFormer showcases its ability to leverage comprehensive data sources and advanced modeling techniques to make accurate predictions in areas where limited data is available.

\section{Summary}
\label{sec:conclusion}
In this study, we have presented $CrashFormer$, a transformer-based network for predicting traffic accidents by leveraging multiple data sources. Our approach combines historical accident and weather data, proximity to points of interest, demographic information, and map images to enhance the accuracy of accident risk prediction. Through rigorous experiments using real-world data from ten major cities in the United States, we have demonstrated the effectiveness of $CrashFormer$ compared to state-of-the-art models. The integration of diverse data sources and the utilization of transformer-based modeling techniques have shown significant improvements in accident risk prediction.



Furthermore, the deployment and evaluation of $CrashFormer$ in real-world settings would be a valuable direction for future research. Conducting large-scale experiments and assessing the model's performance in different cities or regions can provide insights into its practical applicability and scalability.

In conclusion, $CrashFormer$ offers a comprehensive and effective approach to predicting accident risk by leveraging multiple data sources and advanced modeling techniques. The future work outlined above aims to further enhance the model's predictive capabilities and validate its performance in real-world scenarios, ultimately contributing to improved traffic safety and accident prevention efforts.

\bibliographystyle{ACM-Reference-Format}
\bibliography{references}

\end{document}